\documentclass{article}

\PassOptionsToPackage{numbers, compress}{natbib}

\usepackage[final]{neurips_2024}

\usepackage[utf8]{inputenc} 
\usepackage[T1]{fontenc}    
\usepackage{hyperref}       
\usepackage{url}            
\usepackage{booktabs}       
\usepackage{amsfonts}       
\usepackage{amsmath}
\usepackage{graphicx}
\usepackage{nicefrac}       
\usepackage{microtype}      
\usepackage{xcolor}         
\usepackage{multirow}
\usepackage{makecell}
\usepackage{subfigure}

\title{Latent Neural Operator for Solving Forward and Inverse PDE Problems}

\author{%
  Tian Wang \\
  State Key Laboratory of \\
  Multimodal Artificial Intelligence Systems \\
  Institute of Automation, \\
  Chinese Academy of Sciences \\
  School of Artificial Intelligence, \\
  University of Chinese Academy of Sciences \\
  \texttt{wangtian2022@ia.ac.cn} \\
  \And
  Chuang Wang\thanks{Corresponding Author.This work is supported by Beijing Municipal Science and Technology Project (No. Z231100010323005), the 2035 Innovation Mission (Grants No. E4J10102) and  Pioneer Hundred Talents Program of CAS  under Grant Y9S9MS08 and E2S40101.} \\
  State Key Laboratory of \\
  Multimodal Artificial Intelligence Systems \\
  Institute of Automation, \\
  Chinese Academy of Sciences \\
  School of Artificial Intelligence, \\
  University of Chinese Academy of Sciences \\
  \texttt{wangchuang@ia.ac.cn} \\
}

\begin{document}
\global\long\def\vct#1{\boldsymbol{#1}}%
\global\long\def\mat#1{\boldsymbol{#1}}%
\global\long\def\opvec{\text{vec}}%
\global\long\def\optr{\mbox{tr}}%
\global\long\def\opmat{\mbox{mat}}%
\global\long\def\opdiag{\mbox{diag}}%

\global\long\def\t#1{\widetilde{#1}}%
\global\long\def\h#1{\widehat{#1}}%
\global\long\def\abs#1{\left\lvert #1\right\rvert }%
\global\long\def\norm#1{\lVert#1\rVert}%

\global\long\def\inprod#1{\langle#1\rangle}%
\global\long\def\set#1{\left\{  #1\right\}  }%
\global\long\def\bydef{\overset{\text{def}}{=}}%

\global\long\def\EE{\mathbb{E}\,}%
\global\long\def\EEk{\mathbb{E}_{k}\,}%

\global\long\def\R{\mathbb{R}}%
\global\long\def\E{\mathbb{E}}%

\global\long\def\va{\boldsymbol{a}}%
\global\long\def\vb{\boldsymbol{b}}%
\global\long\def\vc{\boldsymbol{c}}%
\global\long\def\vd{\boldsymbol{d}}%
\global\long\def\ve{\boldsymbol{e}}%
\global\long\def\vf{\boldsymbol{f}}%
\global\long\def\vg{\boldsymbol{g}}%

\global\long\def\vh{\boldsymbol{h}}%
\global\long\def\vi{\boldsymbol{i}}%
\global\long\def\vj{\boldsymbol{j}}%
\global\long\def\vk{\boldsymbol{k}}%
\global\long\def\vl{\boldsymbol{l}}%

\global\long\def\vm{\boldsymbol{m}}%
\global\long\def\vn{\boldsymbol{n}}%
\global\long\def\vo{\boldsymbol{o}}%
\global\long\def\vp{\boldsymbol{p}}%
\global\long\def\vq{\boldsymbol{q}}%
\global\long\def\vr{\boldsymbol{r}}%

\global\long\def\vs{\boldsymbol{s}}%
\global\long\def\vt{\boldsymbol{t}}%
\global\long\def\vu{\boldsymbol{u}}%
\global\long\def\vv{\boldsymbol{v}}%
\global\long\def\vw{\boldsymbol{w}}%
\global\long\def\vx{\boldsymbol{x}}%
\global\long\def\vy{\boldsymbol{y}}%
\global\long\def\vz{\boldsymbol{z}}%

\global\long\def\vtheta{\boldsymbol{\theta}}%
\global\long\def\vxi{\boldsymbol{\xi}}%
\global\long\def\vdelta{\boldsymbol{\delta}}%
\global\long\def\veta{\vct{\eta}}%

\global\long\def\mA{\boldsymbol{A}}%
\global\long\def\mB{\boldsymbol{B}}%
\global\long\def\mC{\boldsymbol{C}}%
\global\long\def\mD{\boldsymbol{D}}%
\global\long\def\mE{\boldsymbol{E}}%
\global\long\def\mF{\boldsymbol{F}}%
\global\long\def\mG{\boldsymbol{G}}%

\global\long\def\mH{\boldsymbol{H}}%
\global\long\def\mI{\boldsymbol{I}}%
\global\long\def\mJ{\boldsymbol{J}}%
\global\long\def\mK{\boldsymbol{K}}%
\global\long\def\mL{\boldsymbol{L}}%
\global\long\def\mM{\boldsymbol{M}}%
\global\long\def\mN{\boldsymbol{N}}%

\global\long\def\mO{\boldsymbol{O}}%
\global\long\def\mP{\boldsymbol{P}}%
\global\long\def\mQ{\boldsymbol{Q}}%
\global\long\def\mR{\boldsymbol{R}}%
\global\long\def\mS{\boldsymbol{S}}%
\global\long\def\mT{\boldsymbol{T}}%
\global\long\def\mU{\boldsymbol{U}}%

\global\long\def\mV{\boldsymbol{V}}%
\global\long\def\mW{\boldsymbol{W}}%
\global\long\def\mX{\boldsymbol{X}}%
\global\long\def\mY{\boldsymbol{Y}}%
\global\long\def\mZ{\boldsymbol{Z}}%

\global\long\def\mLa{\boldsymbol{\Lambda}}%
\global\long\def\mOm{\boldsymbol{\Omega}}%

\global\long\def\a{\alpha}%
\global\long\def\b{\beta}%
\global\long\def\m{\mu}%
\global\long\def\n{\nu}%
\global\long\def\g{\gamma}%
\global\long\def\s{\sigma}%
\global\long\def\e{\epsilon}%
\global\long\def\w{\omega}%

\global\long\def\T{\intercal}%
\global\long\def\d{\text{d}}%

\global\long\def\nt{\left\lfloor nt\right\rfloor }%
\global\long\def\ns{\left\lfloor ns\right\rfloor }%
\global\long\def\textif{\text{if }}%
\global\long\def\otherwise{\text{otherwise}}%

\global\long\def\lmax{\lambda_{\max}}%
\global\long\def\lmin{\lambda_{\min}}%

\global\long\def\tvy{\t{\boldsymbol{y}}}%
\global\long\def\tc{\t c}%
\global\long\def\ttau{\t{\tau}}%
\global\long\def\tf{\t f}%
\global\long\def\th{\t h}%
\global\long\def\tq{\t q}%
\global\long\def\tz{\t z}%
\global\long\def\hw{\h w}%
\global\long\def\hvv{\h{\vv}}%
\global\long\def\hvu{\h{\vu}}%

\global\long\def\tva{\t{\boldsymbol{a}}}%
\global\long\def\tvw{\t{\boldsymbol{w}}}%
\global\long\def\tw{\t w}%
\global\long\def\tvc{\t{\vc}}%
\global\long\def\dif{\text{d}}%

\global\long\def\cw#1{{\color{blue} C: #1}}%
\global\long\def\tw#1{{\color{green} M: #1}}%

\maketitle

\begin{abstract}
Neural operators effectively solve PDE problems from data without knowing the explicit equations, which learn the map from the input sequences of observed samples to the predicted values. Most existing works build the model in the original geometric space, leading to high computational costs when the number of sample points is large. We present the Latent Neural Operator (LNO) solving PDEs in the latent space. In particular, we first propose Physics-Cross-Attention (PhCA) transforming representation from the geometric space to the latent space, then learn the operator in the latent space, and finally recover the real-world geometric space via the inverse PhCA map. Our model retains flexibility that can decode values in any position not limited to locations defined in the training set, and therefore can naturally perform interpolation and extrapolation tasks particularly useful for inverse problems. Moreover, the proposed LNO improves both prediction accuracy and computational efficiency. Experiments show that LNO reduces the GPU memory by 50\%, speeds up training 1.8 times, and reaches state-of-the-art accuracy on four out of six benchmarks for forward problems and a benchmark for inverse problem. {Code is available at \href{https://github.com/L-I-M-I-T/LatentNeuralOperator}{https://github.com/L-I-M-I-T/LatentNeuralOperator}.}

\end{abstract}

\section{Introduction}
Neural network approaches for partial differential equations (PDEs) attract increasing attention in diverse scientific fields, for instance, meteorological prediction\cite{Fourcastnet}, industrial design\cite{ComplexStress}, geological sensing\cite{DeepONetForwardInverse} and environmental monitoring\cite{FINN_GroundwaterContaminant} to name a few. 
Compared with traditional numerical methods, {\em e.g.}, finite element method\cite{FiniteElementMethods}, which demands a large number of computational resources and requires meticulous discretization involving a lot of specialized knowledge, neural networks provide a new paradigm for solving PDE-driven problems with a better trade-off between accuracy and flexibility.

Neural PDE solvers can be divided into two main categories, physics-informed and operator-learning models. The physics-informed methods\cite{PIML,PINN,FINN,DeepGreen,PeRCNN,PINN_DeepONet} enforce the inductive bias of explicit PDE instances into loss function or model architecture, usually yielding solutions with high accuracy, but are less flexible as they require fresh training to find the solution for each new instance. 
In contrast, the operator learning methods\cite{ChenChenTheroem,DeepONet,FNO,NO} adopt a data-driven paradigm to learn the implicit PDE constraint from pair of samples representing input-to-output functional mappings, providing better flexibility that can generalize to different initial/boundary conditions beyond the scope of the training data, but have yet large room to improve in the prediction accuracy.

Recently, Transformers\cite{Attention} prevail in neural operator structures. Theoretically, the attention mechanism was proved to be a special form of the integral kernel\cite{NO,LOCA} in neural operator, which naturally conforms to sequence-to-sequence characterization. Moreover, the fully-connected attention structure enables the model to characterize long-distance and quadratic relationships among sample points, which yields more accurate prediction than the vanilla MLP structure\cite{DeepONet} but at the expense of increasing the computational complexity drastically, as the complexity is quadratic respect to the length of input sequence.

To alleviate the complexity, existing works employed linear attention mechanism \cite{EA,NA,GalerkinTransformer,GNOT} speeding up the computation but sacrificing precision due to their limited expressive power. Another line of works attempted to solve the PDE in the latent space with a small number of samples \cite{LSM,Transolver,UPT,LST,LA}, which get rid of abundant sample points in the original geometric space and capture physical correlations in the tighter latent space.

In this work, we present the Latent Neural Operator (LNO) with particular effort on designing the \textbf{Ph}ysics-\textbf{C}ross-\textbf{A}ttension (PhCA) module to learn the latent space to optimize the trade-off between accuracy and flexibility, and reduce the computational complexity as well. 
Specifically, LNO first encodes the input sample sequence into a learnable latent space by PhCA. Then, the model learns the PDE operator in the latent space via a stack of Transformer layers. Finally, the predicted output function is decoded by the inverse PhCA, given any query locations. 
Unlike previous works that either pre-define the latent space as frequency domain \cite{FNO}, or switch back-and-forth between latent space and geometric space at each layer \cite{Transolver}, the latent space of the LNO is learnable from data and the data are only transformed at the first and last layer. 

The proposed PhCA module decouples the locations of input samples fed in the encoder and output samples fed in the decoder, which can predict values in any position not limited by the observation. Therefore, it can naturally perform interpolation and extrapolation tasks, which is fundamental for solving inverse problems.
The proposed LNO improves both prediction accuracy and computational efficiency. Experiments show that LNO reduces the GPU memory by 50\% and speeds up training 1.8 times. Moreover, it reaches state-of-the-art accuracy on four (Pipe, Elasticity, Darcy, Navier-Stokes) out of six benchmarks for forward problems and a benchmark for inverse problem on Burgers' equation.

The main contributions of our LNO framework are summarized as follows. 
\begin{itemize}
\item Flexibility: We propose the Physics-Cross-Attention (PhCA) module which decouples the locations of input and output samples and learns the latent space from data. We also build the Latent Neural Operator (LNO) model for solving both forward and inverse PDE problems.
\item Efficiency: The LNO model transforms the data to the latent space only once. Compared with other approaches that switch the latent space and geometric space in each Transformer layer, the proposed model reduces complexity drastically in terms of GPU memory usage, training time and number of model parameters.
\item Accuracy: We obtain state-of-the-art results on the forward problems of Pipe, Elasticity, Darcy, Navier-Stokes, and the inverse problem of Burgers' equation.
\end{itemize}

\section{Related works}
We first review two major classes of neural PDE approaches, physics-informed and operator-learning methods. Then, we briefly discuss the works using neural networks for solving inverse PDE problems.

\subsection{Physics-Informed Machine Learning}
Physics-informed machine learning methods incorporate prior knowledge of PDEs into loss function or model architecture. The Physics-Informed Neural Network (PINN)\cite{PINN} and its variants 
are among the most classic methods. PINN integrates the differential equation, initial conditions and boundary conditions into the loss function, guiding the neural network to approximate the solution by ensuring its derivatives satisfies the equation and the output aligns with the initial and boundary conditions. Other methods attempt to encode physical priors into the model architecture, {\em e.g.,} the FInite volume Neural Network (FINN)\cite{FINN} and Physics-encoded  Recurrent Convolutional Neural Network (PeRCNN)\cite{PeRCNN}. These methods achieve high precision but require solving optimization to train the networks for each instance. In addition, in the case, where concrete PDE is unknown or not exact, it is challenging to adopt those methods, limiting the generalization ability and flexibility in practical data-rich situations.

\subsection{Operator Learning}
Operator learning methods aim to find a functional map from coefficients, forcing terms and initial conditions to the solution function by learning an approximating operator. These methods train neural networks to capture the correspondence between input and output functions from data without needing to know the underlying PDE explicitly. DeepONet\cite{DeepONet,DeepONetTL} first introduces approximating operators using neural networks. 

A series of kernel integral operator forms were proposed to model the mappings between functions as neural operators\cite{GNO,FNO,NO}. 
Specifically, Galerkin Transformer\cite{GalerkinTransformer} utilizes the Galerkin-type attention mechanism as kernel integral operators.  OFormer\cite{OFormer} improves Galerkin-type attention by decoupling observation positions and query positions through cross-attention. GNOT\cite{GNOT} introduces heterogeneous normalized cross-attention to modulate various input conditions for query positions and alleviates computational pressure of the large grid by linear-time-complexity attention mechanism. FactFormer\cite{FactFormer} proposes axial factorized kernel integral, decomposing input functions into multiple one-dimensional sub-functions, effectively reducing computation. ONO\cite{ONO} enhances the generalization of neural operators by adding orthogonality regularization in attention mechanisms. 

Researchers also explored the idea of modeling in the latent space. Transolver\cite{Transolver} employs Physics-Attention in each layer to map back-and-forth between geometric features and physical features. LSM\cite{LSM} encodes the input function into the latent space using cross-attention, constructs a set of orthogonal bases, and then decodes the solution back to the geometric space through cross-attention again. UPT\cite{UPT} introduces encoding and decoding losses, and compresses geometric point information into super-node information in the latent space via graph neural network\cite{GNN}.

Compared with existing methods, we autonomously learn mappings between functions in the real-world space and the latent space in an end-to-end fashion with the cross-attention mechanism, without manually constructed orthogonal basis nor additionally introduced loss.

\subsection{Inverse Problem}
Inverse problems of PDEs have extensive applications in fields involving sensing and reconstruction such as medical imaging and geological sensing. For instance, Transformer-based deep direct sampling methods\cite{TransformerBoundaryInverseProblem} have been proposed for electrical impedance tomography, and convolutional network methods\cite{ConvolutionalAutoEncoder,GPRInvNet,VelocityGAN} are widely used in full waveform inversion. 

Theoretical works on solving inverse problems have also been proposed. The reconstruction of diffusion fields with localized sources has been studied\cite{LuSampling} from the perspective of sampling matrix in the spatiotemporal domain. PINN\cite{PINN_Inverse1,PINN_Inverse2,PINN_Inverse3} approximates the solution based on partially observed data to compute the unknown data. NIO\cite{NIO} combines DeepONet\cite{DeepONet} and FNO\cite{FNO} to construct a framework for solving inverse problems. FINN\cite{FINN,FINN_BC} sets boundary conditions as learnable parameters, and infers boundary values through backpropagation. The generative method\cite{LSI} generates the initial conditions through latent codes and refines them based on the results of time evolution. We aim to unify the solution of forward and inverse problems in the latent space through the PhCA module, which decouples observation and prediction positions. 

\section{Method}
We formally define both the forward and the inverse problems of PDEs. Then, we introduce our Latent Neural Operator (LNO) model. Finally, we discuss the core module that learns the transformation between the real-world geometric space and the latent space via the cross-attention mechanism.

\begin{figure}
\centering
\includegraphics[width=1.0\textwidth]{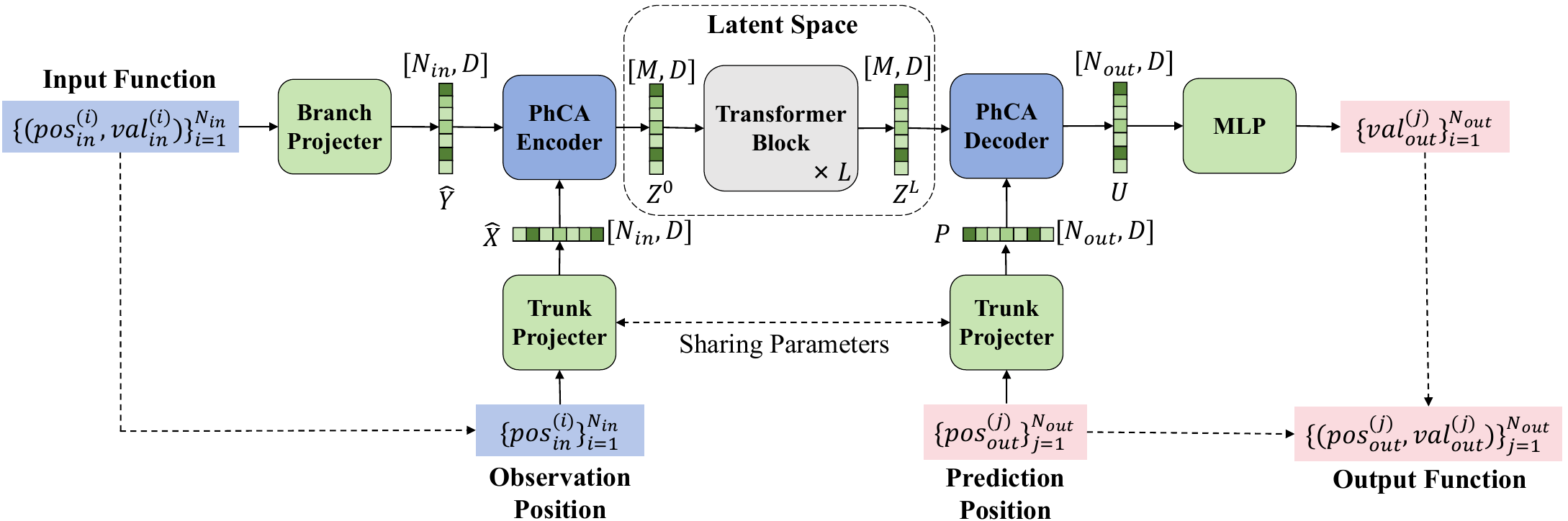}
\caption{The overall architecture of Latent Neural Operator.} 
\label{fig-1}
\end{figure}

\vspace{-5pt}
\subsection{Problem Setup}
We consider a partial differential equation (PDE) with boundary and/or initial conditions defined on $\Omega \subseteq \R^d$ or $\R^d \times [0,T)$ 
\begin{equation}\label{eq:PDE} 
  \begin{aligned}
    \mathcal{L}_{a}\circ u&=0, \\
    u({x})&=b({x}),\quad {x}\in \partial \Omega 
  \end{aligned}
\end{equation}
where $\mathcal{L}_{a}$ is an operator containing partial differentials parameterized by a set of coefficients $a$, and $u(x)$, $x \in \Omega$ and $b({x})$ represent the solution to the PDE and the boundary/initial conditions respectively.

A conventional forward problem aims to find the solution $u(x)$, given the differential operator $\mathcal{L}_a$ in  \eqref{eq:PDE} with the explicit parameters $a$ and the boundary/initial conditions $b(x)$.
%
For an inverse problem, given an observed set of partial data  $\{(x, u(x))|\;x \in \tilde{\Omega}\subset \Omega\}$, the task is to infer the PDE that generates the observed data along with possible boundary/initial conditions or the parameters $a$ in the differential operator $\mathcal{L}_a$. 

Both forward and inverse problems can be unified as an operator learning task fitting into a sequence-to-sequence translation framework. An operator model estimates the mapping $\mathcal{F}: f \mapsto g$ from an input function $f$ to an output function $g$ by the data-driven approach using paired training data of input-output sequences
$( \{f(y_i)\}_i,\{g(z_j)\}_j)$, where $y_i$ and $z_j$ correspond to positions in the domain of the input and output function respectively. 

Specifically, for the forward problem, the inputs are samples of either boundary/initial function $b$ or PDE coefficient function $a$, and the output is the physical quantity $u$ evaluated at a grid of points.  Conversely, for the inverse problem, the input is a few partially observation samples $\{(pos_{in}^{(i)},val_{in}^{(i)})\}^{N_{in}}_{i=1}=\{ (x_i, u(x_i) ) \}_{i=1}^{N_{in}}$, and the outputs are boundary/initial function ({\em e.g.,} infer initial state) or the PDE coefficient function (also coined as system identification problem). 

The challenge in solving forward and inverse problems of PDEs lies in the large lengths of input and output sequences. In terms of operator approximating, complex kernel integrals lead to high computational cost while simple kernel integrals lack accuracy. In this work, we build an accurate and computationally efficient model for both forward and inverse problems by learning the operator in the learnable latent space.

\subsection{Latent Neural Operator}
\label{LNO}
\paragraph{Overview}  The proposed model of Latent Neural Operator (LNO) is applicable for both forward and inverse PDE problems. It consists of four modules: an embedding layer to lift the dimension of the input data, an encoder to transform the input data to a learnable latent space, a series of self-attention layers for modeling operator in the latent space, and a decoder to recover the latent representation to the real-world geometric space. The overall architecture of LNO is shown in Figure~\ref{fig-1}.

\paragraph{Embedding inputs}
The geometric space is the original space of the PDE input or output which contains $N_{in}$ or $N_{out}$ samples, each with $d$-dimensional position coordinates (which may additionally include a $1$-dimensional time coordinate for time-dependent PDEs) and $n$-dimensional physical quantity values. 

First, we embed the positions and physics quantity values of the input sequence in the geometric space to higher $D$-dimensional representations. This process is implemented by two MLPs, where the first only embeds the position information and the second embeds both the location and physics quantity information
\begin{equation}\label{eq:embed}
    \begin{aligned} 
    {\hat{x}}^{(i)} &= {\rm trunk\text{-}projector}
    \big({pos}_{in}^{(i)}
    \big), 
    &{\hat{x}}^{(i)}\in \mathbb{R}^{D \times 1} \nonumber
    \\
    {\hat{y}}^{(i)} &= {\rm branch\text{-}projector}
    \big({\rm concat}({pos}_{in}^{(i)},
    {val}_{in}^{(i)})
    \big), \quad&{\hat{y}}^{(i)}\in \mathbb{R}^{D \times 1}. \nonumber
    \end{aligned}
\end{equation}
Analogous to DeepONet \cite{DeepONet}, we name these MLPs as trunk-projector and branch-projector respectively. The embedding operation helps to map the locations and physics quantities to the embedding space where their relationships are easier to capture.

Removing the physics quantity values from the trunk-projector input grants our model the decoupling property. In this way, we can predict values at positions not included in the sampled input function from the latent space, thus achieving the decoupling of observation and prediction positions. Since the sampled input function always includes position-value pairs, using both position and value information in the branch projector will not affect the decoupling objective.

\paragraph{Encoding to the latent space}
We attempt to represent the input function using the representation tokens of $M$ hypothetical sampling positions $\{h^{(k)}\}^{M}_{k=1},h^{(k)}\in R^{D\times 1}$ which exist in the latent space, where $M$ can be much smaller than the number $N_{in}$ of samples of raw inputs.

We use the Physics-Cross-Attention to model the transformation from the geometric space to the latent space. The inputs are query $\mH=[h^{(1)},h^{(2)}...,h^{(M)}]^T$, key $\hat{\mX}=[\hat{x}^{(1)}, \hat{x}^{(2)},\ldots,\hat{x}^{(N_{in})}]^T$ and value $\hat{\mY}=[\hat{y}^{(1)}, \hat{y}^{(2)},\ldots,\hat{y}^{(N_{in})}]^T$ which corresponding to embedding of hypothetical locations in the target latent space, embedding of locations in the source real-world geometric space, and embedding of the concatenation of locations and physics quantities in the source real-world geometric space respectively. The output can be calculated through
\begin{equation} \label{eq:encoder}
\boldsymbol{Z}^0 = {\rm Physics\text{-}Cross\text{-}Attention}(\boldsymbol{H}, \boldsymbol{\hat{X}}, \boldsymbol{\hat{Y}})
\end{equation}
In contrast to FNO\cite{FNO}, where the latent space is pre-defined as the frequency domain, we learn the latent space from data. Therefore, the embeddings of sample locations $\mH$ as queries in the cross-attention are also learnable parameters. 

After encoding the input function into the latent space, the length of the sequence to be processed is reduced from $N_{in}$ to $M$. With  $M\ll N_{in}$, extracting and converting the feature of the input function in the latent space will be much more efficient than in the real-world geometric space. 

\paragraph{Learning operator in the latent space}
We model the operator of the forward/inverse PDE problem in the latent space to convert the feature of the input function to that of the output function, using a stack of Transformer blocks with the self-attention mechanism
\begin{gather}
\boldsymbol{Z}^{l} = {\rm MLP}({\rm LayerNorm}(\boldsymbol{\hat{Z}}^{l})) + \boldsymbol{\hat{Z}}^{l}
\text{ \quad with } \boldsymbol{\hat{Z}}^{l} = {\rm Attention}({\rm LayerNorm}(\boldsymbol{Z}^{l-1})) + \boldsymbol{Z}^{l-1} \nonumber
\end{gather}
where $l\in\{1,2,...,L\}$ is the index of Transformer block with $L$ being the total number of the blocks. We experiment with several implementations of attention mechanisms in the ablation study and find the classical scaled dot-product attention\cite{Attention} performs consistently well on various tasks. 

\paragraph{Decoding to the geometric space}
Finally, we decode the output sequence from the latent sequence through Physics-Cross-Attention according to the query ${pos}_{out}$ of the output sampling locations and use another MLP to map the embeddings to the values of the output function
\begin{equation} \label{eq:decoder}
\begin{aligned}
{p}^{(j)} &= {\rm trunk\text{-}projector}({pos}_{out}^{(j)}), &{p}^{(j)}\in \mathbb{R}^{D \times 1}
\\
\boldsymbol{U}&={\rm Physics\text{-}Cross\text{-}Attention}(\boldsymbol{P}, \boldsymbol{H}, \boldsymbol{Z}^L) 
\\ 
{{val}}_{out}^{(j)}&={\rm MLP}({u}^{(j)}), &{u}^{(j)}\in \mathbb{R}^{D \times 1} 
\end{aligned}
\end{equation}
where $\mP=[p^{(1)}, p^{(2)},\ldots,p^{(N_{out})}]^T$ and $\mU=[u^{(1)}, u^{(2)},\ldots,u^{(N_{out})}]^T$. During the inferring phase, the trunk-projector can accept inputs from any position, enabling LNO to generalize to the unseen region where the positions are not presented in the training phase. 

\vspace{-5pt}
\subsection{Physics-Cross-Attention}
\label{PhCA}
The \textbf {Ph}ysics-\textbf{C}ross-\textbf{A}ttension (PhCA) is the core module in the aforementioned latent neural operator (LNO) model, acting as the encoder and decoder which transform the data representation back-and-forth between the real-world geometric space and the latent space. Motivated by the work \cite{LSM} that learns the operator mapping across domains using the cross-attention, we use the same cross-attention to model the transformation to/from the latent space but set the latent space learnable rather than pre-defined, {\em e.g., } triangular space or frequency domain \cite{FNO}.

In particular, the rows of the query $\mH$ and key $\hat{\mX}$ matrices in \eqref{eq:encoder} represent the embedded positions of samples in the target and source space respectively. A row of the value matrix $\hat{\mY}$ is the embedding of both the position and physics quantity of a sample in the source space. We design the latent space to be learned from data. Thus, the positions $\mH$ of samples in the latent space should be set as learnable parameters as well. Then, the cross-attention of the encoder in \eqref{eq:encoder} is simplified as
\begin{equation} \label{eq:encoder-sim}
\boldsymbol{Z}^0  ={\rm softmax}\Big(\tfrac{1}{\sqrt{D}}\boldsymbol{H}\boldsymbol{W}_{q}\boldsymbol{W}_k^T\boldsymbol{\hat{X}}^T
\Big)
\boldsymbol{\hat{Y}}\boldsymbol{W}_v
={\rm softmax}\big(\boldsymbol{W}_1\boldsymbol{\hat{X}}^T \big)\boldsymbol{\hat{Y}}\boldsymbol{W}_v,
\end{equation}
where we merge the learnable matrices of query $\mH$, query weight $\mW_q$ and key weight $\mW_k$ as a single matrix $\mW_1$.

\begin{figure}[tp]
\centering
\includegraphics[width=0.8\textwidth]{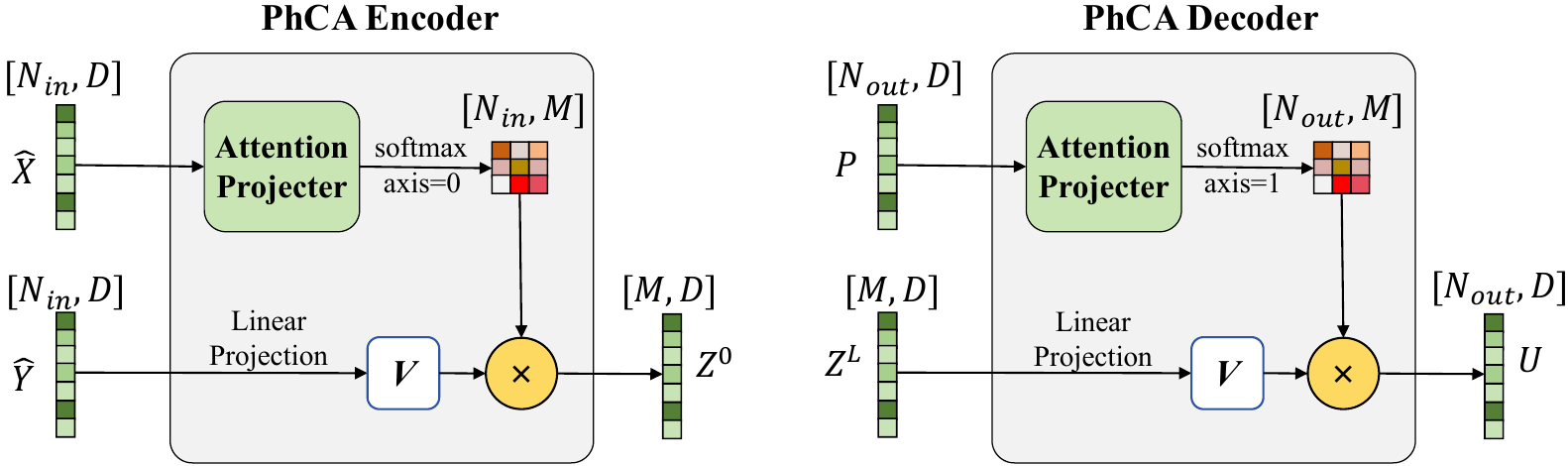}
\caption{The working mechanism of Physics-Cross-Attention in encoder and decoder respectively.} 
\label{fig-2}
\end{figure}

Correspondingly, the cross-attention in the decoder has a learnable key matrix that represents positions of samples in the latent space. Therefore, the cross-attention of the decoder in ~\eqref{eq:decoder} is written as
\begin{equation} \label{eq:decoder-sim}
 \boldsymbol{U}={\rm softmax}\Big(
 \tfrac{1}{\sqrt{D}} \boldsymbol{P}\boldsymbol{W}_q^{'}\boldsymbol{W}_k^{'T}\boldsymbol{H}^T
 \Big)
 \boldsymbol{Z}^L\boldsymbol{W}_v^{'}={\rm softmax}(\boldsymbol{P}\boldsymbol{W}_2^T)\boldsymbol{Z}^L\boldsymbol{W}_v^{'}.
\end{equation}
In addition, utilizing the relationship of the encoder and decoder as mutually inverse transformations, we set $\mW_1=\mW_2$ to reduce the number of parameters, which also improves model performance experimentally. Practically, we generalize the linear projection $\mW_1,\mW_2$ with an MLP for further improvement, which is named attention-projector as shown in Figure~\ref{fig-2}.

In the special case where the input only contains sample positions without physics quantities, our PhCA module is similar to the transformation of physics-aware token in Transolver \cite{Transolver} (only differing in normalization). In general cases when the inputs include both positions and physics quantities, the two approaches are different.
Specifically, in Transolver, the key $\hat{\mX}$ in \eqref{eq:encoder-sim} and query $\mP$ in \eqref{eq:decoder-sim} contain both positions and physics quantities information, whereas in PhCA, the key matrix $\hat{\mX}$ in the encoder and the query matrix $\mP$ in the decoder only depend on the positions of the input and output respectively. The physics quantities are used only to produce the value matrix $\mV$ in attention mechanism. Such decoupling property of values and locations in query and key enables LNO to predict values at different positions from input positions.

\vspace{-5pt}
\section{Experiments}
We first evaluate the accuracy for the forward problems on the six popular public benchmarks and for the inverse problem using Burger's equation.
Then, we measure the efficiency in terms of the number of model parameters, the GPU memory usage and the training time.
Next, we assess the generalization of our model concerning the number $N_{in} $ of observation samples and the number $N_{out}$ of prediction positions.
Finally, we establish a series of ablation studies including choices of different attention mechanisms, strategy of sharing weights and sampling numbers in the latent space. Model is developed on the PyTorch and the PaddlePaddle deep learning framework. Additional experiments on scaling, sampling rate, and significance are presented in the appendix.

\subsection{Accuracy for the forward problems}
\label{forward-exp}
We conduct experiments for the forward problem on six benchmarks including Darcy and NS2d on regular grids (refer to section 5.2, 5.3 in \cite{FNO} for details), and Airfoil, Elasticity, Plasticity and Pipe problems on irregular grids (refer to section 4.1--4.5 in \cite{GeoFNO} for details).

Table \ref{tab-1} shows a comprehensive comparison with previous works that reported their performance on the same benchmarks. Our LNO achieves the best accuracy on four benchmarks: Darcy, NS2d, Elasticity and Pipe. On the Airfoil and Plasticity benchmarks, our LNO achieves performance close to the SOTA with only half computational cost as discussed in Section \ref{Efficiency}. These results demonstrate the effectiveness of approximating the kernel integral operator in the latent space after transformation from the real-world geometric space through the PhCA module.

\begin{table}[htbp]
\centering
\caption{Prediction error on the six forward problems. The best result is in bold. "$\ast$" means that the results of the method are reproduced by ourselves. "$/$" means that the method can not handle this benchmark. The column labeled D.C. indicates whether the observation positions and prediction positions are decoupled.}
\label{tab-1}
\begin{tabular}{lccccccc}
\toprule
\multirow{2}*{Model}&\multirow{2}*{D.C.}&\multicolumn{6}{c}{Relative L2($\times 10^{-2}$)} \\
\cmidrule(r){3-8}
~&~& Darcy & NS2d & Airfoil & Elasticity & Plasticity & Pipe\\
\midrule
FNO\cite{FNO}&N&1.08&15.56&/&/&/&/ \\
Geo-FNO\cite{GeoFNO}&N&1.08&15.56&1.38&2.29&0.74&0.67\\
F-FNO$^{\ast}$\cite{FFNO}&N&0.75&11.51&0.60&1.85&0.27&0.68\\
U-FNO$^{\ast}$\cite{UFNO}&N&1.28&17.15&1.19&2.13&0.41&0.62\\
LSM\cite{LSM}&N&0.65&15.35&0.59&2.18&0.25&0.50\\
\cmidrule(r){0-7}
Galerkin\cite{GalerkinTransformer}&N&0.84&14.01&1.18&2.40&1.20 &0.98\\
OFormer\cite{OFormer}&Y&1.24&17.05&1.83&1.83&0.17&1.68\\
GNOT$^{\ast}$\cite{GNOT}&Y&1.04&13.40&0.75&0.88&3.19&0.45\\
FactFormer\cite{FactFormer}&N&1.09&12.14&0.71&/&3.12&0.60 \\
ONO\cite{ONO}&Y&0.76&11.95&0.61&1.18&0.48&0.52\\
Transolver$^{\ast}$\cite{Transolver}&N&0.58&8.79&\textbf{0.47}&0.62&\textbf{0.12}&0.31 \\
LNO(Ours)&Y&\textbf{0.49}&\textbf{8.45}&0.51&\textbf{0.52}&0.29&\textbf{0.26} \\
\bottomrule
\end{tabular}
\end{table}
\vspace{-5pt}
\subsection{Accuracy for the inverse problem}
\label{inverse-exp}
To demonstrate the flexibility of LNO, we design an inverse problem. Given the partially observed solution $u(x)$, the aim is to recover the complete $u(x)$ in a larger domain. 
Specifically, we test on  1D Burgers' equation
\begin{gather*} 
\tfrac{\partial }{\partial t}u(x,t) = 0.01\tfrac{\partial^2 }{\partial x^2} u(x,t)-u(x,t)\tfrac{\partial }{\partial x} u(x,t), \quad x\in[0,1],\; t\in[0,1] \\
u(x,0)\sim \text{GaussianProcess}(0, \exp\big(-\tfrac{2}{pl^2}\sin^2(\pi ||x-x^{'}||\big), \quad u(0, t)=u(1, t)
\end{gather*}
The ground-truth data is generated on a $128\times128$ grid with the periodic boundary condition. Initial conditions are sampled from the Gaussian process with the periodic length $p=1$, scaling factor $l=1$. 

The objective of this inverse problem is to reconstruct the complete solution $u(x)$  in the whole spatiotemporal domain $(x,t)\in[0,1]\times[0,1]$ based on sparsely random-sampled or fixed-sampled observation in the sub-spatiotemporal domain $(x,t)\in[0,1]\times[0.25,0.75]$, as illustrated in Figure~\ref{fig-3} in the appendix. 

Instead of the naive approach which directly learns the mapping from partially observed samples in the subdomain to the complete solution in the whole domain, we propose a two-stage strategy inspired by inpainting\cite{inpainting1,inpainting2} and outpainting\cite{outpainting1,outpainting2}. First, we train the model as a completer to interpolate the sparsely sampled points in the subdomain $[0,1]\times[0.25,0.75]$ (left sub-figure in Figure \ref{fig-3}) to predict all densely and regularly sampled points in the same subdomain (middle sub-figure in Figure \ref{fig-3}). Then, we train the model as a propagator to extrapolate the results of the completer from the subdomain to the whole domain $[0,1]\times[0,1]$ (right sub-figure in Figure \ref{fig-3}). Since the observation and the prediction samples are located in different positions, only the models with decoupling properties can be used as the completer and propagator. 

We compare the performance of our LNO with DeepONet\cite{DeepONet} and GNOT\cite{GNOT} in two stages respectively. The results in Table~\ref{tab-2} and Table~\ref{tab-3} indicate that i) in the first stage, LNO performs significantly better than DeepONet and GNOT at different random observation ratios; ii) in the second stage, LNO performs better when the complete solution in the subdomain approaches the ground truth, and the performance degrades as the reconstruction error of the complete solution in the subdomain increases. This is reasonable since we train the propagator based on the ground truth in the subdomain. When there is a significant discrepancy between the ground truth and the prediction of the completer, the model faces challenging distribution shifts. Nevertheless, when LNO is used both as the completer and propagator, it achieves the most accurate results for solving this inverse problem. 

We also investigate the impact of different temporal and spatial sampling intervals on the accuracy of LNO  in the fixed grid observation situation. See appendix \ref{appendix-B} for details.

\begin{table}[htbp]
\centering
\caption{Relative MAE of different completers in the subdomain in the 1st stage of the inverse problem with different settings of observation ratio. }
\label{tab-2}
\begin{tabular}{lccccc}
\toprule
Completer $|$ Observation ratio &20\%&10\%&5\%&1\%&0.5\%  \\
\midrule
DeepONet\cite{DeepONet}&2.51\%&2.59\%&2.82\%&3.25\%&4.82\%\\
GNOT\cite{GNOT}&1.12\%&1.39\%&1.62\%&1.63\%&2.56\%\\
LNO(Ours)&\textbf{0.60\%}&\textbf{0.74\%}&\textbf{0.77\%}&\textbf{1.18\%}&\textbf{2.05\%}\\
\bottomrule
\end{tabular}
\end{table}

\begin{table}[htbp]
\centering
\caption{The reconstruction error of different propagators in the 2nd stage of the inverse problem. Propagators are trained to reconstruct the solution in the whole domain based on the ground truth (G.T.) of the subdomain and are tested using the output of different completers. Relative MAE of $t=0$ and $t=1$ is reported.}
\label{tab-3}
\begin{tabular}{lccc|cc|cc}
\toprule
\multirow{2}*{Propagator $|$ Completer }&\multirow{2}*{G.T.}&\multicolumn{2}{c}{LNO}&\multicolumn{2}{|c}{GNOT}&\multicolumn{2}{|c}{DeepONet}  \\
\cmidrule{3-8}
~&~&10\%&1\%&10\%&1\%&10\%&1\%\\
\midrule
DeepONet\cite{DeepONet}&7.34\%&8.01\%&9.38\%&9.09\%&10.80\%&11.14\%&13.87\%\\
GNOT\cite{GNOT}&5.45\%&6.50\%&8.07\%&\textbf{8.04\%}&\textbf{9.91\%}&\textbf{10.41\%}&\textbf{13.45\%}\\
LNO(Ours)&\textbf{3.73\%}&\textbf{5.69\%}&\textbf{7.72\%}&9.03\%&10.98\%&13.11\%&15.50\%\\
\bottomrule
\end{tabular}
\end{table}

\begin{table}[htbp]
\centering
\caption{Comparison of efficiency between LNO and Transolver on the six forward problem benchmarks. Three metrics are measured: the number of parameters, the cost of GPU memory and the cost of training time per epoch.}
\label{tab-5}
\begin{tabular}{clccccccc}
\toprule
Metric&Model& Darcy & NS2d & Airfoil & Elasticity & Plasticity & Pipe\\
\midrule
\multirow{2}*{Paras Count(M)}&Transolver&1.91&5.33&1.91&1.91&1.91&1.91\\
~&LNO&\textbf{0.76}&\textbf{5.08}&\textbf{1.36}&\textbf{1.42}&\textbf{1.36}&\textbf{1.36}\\
\midrule
\multirow{2}*{Memory(GB)}&Transolver&17.11&17.17&4.49&1.48&18.41&5.94\\
~&LNO&\textbf{5.75}&\textbf{7.58}&\textbf{2.47}&\textbf{1.39}&\textbf{7.16}&\textbf{2.89}\\
\midrule
\multirow{2}*{Time(s/epoch)}&Transolver&88.68&107.62&19.49&5.66&83.43&25.56\\
~&LNO&\textbf{38.98}&\textbf{57.83}&\textbf{9.35}&\textbf{5.33}&\textbf{41.62}&\textbf{14.13}\\
\bottomrule
\end{tabular}
\end{table}

\subsection{Efficiency}
\label{Efficiency}
We demonstrate improvement in the efficiency of our proposed LNO compared to Transolver, which was considered the most efficient Transformer-based PDE solvers\cite{Transolver}. Results in Table~\ref{tab-5} indicate that our method has fewer parameters, lower memory consumption, and faster training speed. Roughly speaking, LNO reduces the parameter count by an average of 30\% and memory consumption by 50\% compared to Transolver across all benchmarks, also with a 1.8x training acceleration. 

The improvement of LNO stems from the idea of solving PDEs in the latent space. Given $N$ input points in the real-world geometric space, if both Transolver and LNO consist of $L$ Transformer blocks, the Transolver, which applies Physics-Attention in each block, has the time complexity of $O(LMN+LM^2)$, where $M$ is the number of slices. Our LNO, which applies Physics-Cross-Attention only in the encoder and decoder, has the time complexity of $O(MN+LM^2)$, where $M$ is the sampling numbers in latent space.

\subsection{Generalization}
\label{Generalization}
To validate the generalization of our model concerning the number of discretization points in the input and output, we train the model in a uniform low-resolution scenario and perform zero-shot inference in other higher-resolution scenarios. Specifically, we downsample the Darcy dataset with a resolution of $421\times421$ to $241\times241$, $211\times211$, $141\times141$, $85\times85$, $61\times61$ and $43\times43$. Models are trained on the $43\times43$ dataset and tested on the other datasets. The comparison results of our model and other methods shown in Table ~\ref{tab-2e} show that our model exhibits great generalization capability regarding the number of sampling points, consistently outperforming other methods in accuracy on test sets with a different number of discretization points from the training set.

\begin{table}[htbp]
\centering
\caption{Prediction errors on Darcy with different resolutions. Models are trained on $43\times43$ and tested on higher resolutions. Relative L2 errors are reported. The best results are in bold.}
\label{tab-2e}
\begin{tabular}{lccccc}
\toprule
Model&$61\times61$&$85\times85$&$141\times141$&$211\times211$&$241\times241$\\
\midrule
FNO&0.1164&0.1797&0.2679&0.3160 &0.3631\\
ONO&0.0204&0.0259&0.0315&0.0349&0.0386\\
GNOT&0.0256&0.0275&0.0294&0.0305&0.0317\\
Transolver&0.0239 &0.0240 &0.0258&0.0279&0.0290\\
LNO(Ours)&\textbf{0.0179}&\textbf{0.0192}&\textbf{0.0214}&\textbf{0.0229}&\textbf{0.0244}\\
\bottomrule
\end{tabular}
\end{table}

\subsection{Ablation Study}
\label{Ablation}
\paragraph{Necessity of decoupling}
Decoupling the values and locations as well as the observation positions and prediction positions not only guarantees flexibility but also improves accuracy. To study the effect of decoupling property, we replace the PhCA in LNO with Physics-Attention proposed in Transolver\cite{Transolver}. In this way, the query and keys are calculated from the same embedding of both ${pos}_{in}$ and ${val}_{in}$ of the input sequence. The ${pos}_{out}$ which is not present in the input sequence can not generate queries and keys. So the observation positions and prediction positions must be kept the same. We compare the performance of LNO with PhCA or Physics-Attention on the forward problem benchmarks, and the results in the first and last rows of  Table~\ref{tab-4} verify the necessity of decoupling.

\paragraph{Expressiveness of quadratic attention}
We study the impact on accuracy using different attention mechanisms in the latent space.
We replace the scaled dot-product attention in LNO with Galerkin-type attention used in Galerkin Transformer\cite{GalerkinTransformer}, efficient attention\cite{EA} used in GNOT\cite{GNOT}, and Nystrom attention\cite{NA} used in ONO\cite{ONO}. 
The comparison results in Table~\ref{tab-4} show that quadratic attention, {\em i.e.,} the scaled dot-product attention\cite{Attention}, is more expressive than linear attention, which helps improve the accuracy. Many previous methods have employed different linear attentions\cite{GalerkinTransformer, GNOT, ONO} to reduce the computational costs at the expense of reducing the accuracy. Since the length of the latent sequence is much smaller than the inputs and outputs in real-world geometric space, applying quadratic attention in latent space does not incur significant overhead. 

\paragraph{Sharing parameters}
We explore the performance of LNO when ${W}_1 \ne {W}_2$ in \eqref{eq:encoder-sim} and \eqref{eq:decoder-sim}. In this way, we replace the linear projection $W_1$ and $W_2$ with two individual MLPs. Results of the comparison between LNO using shared MLPs and non-shared MLPs in the PhCA encoder and decoder are shown in Table~\ref{tab-3e}. It can be found that LNO performs better on most benchmarks (except Plasticity) under the shared situation.

\begin{figure}[htbp]
\centering
\subfigure[]{
\label{fig-4a}
\includegraphics[width=0.31\linewidth]{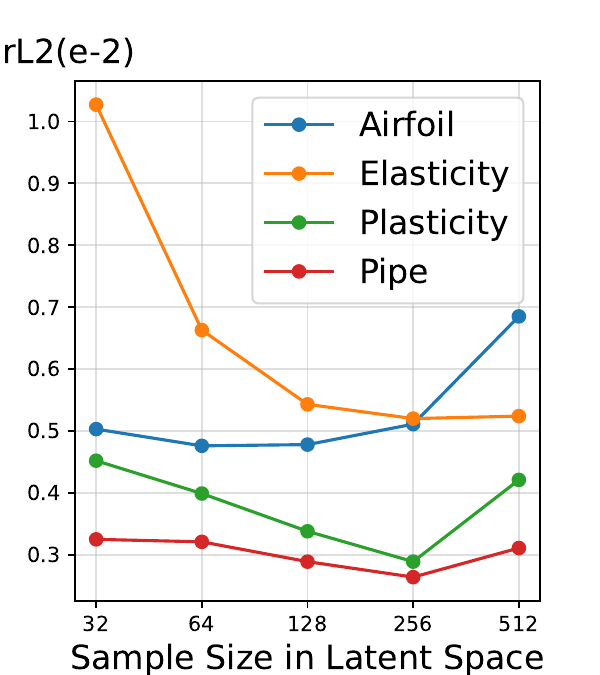}
}
\noindent
\subfigure[]{
\label{fig-4b}
\includegraphics[width=0.31\linewidth]{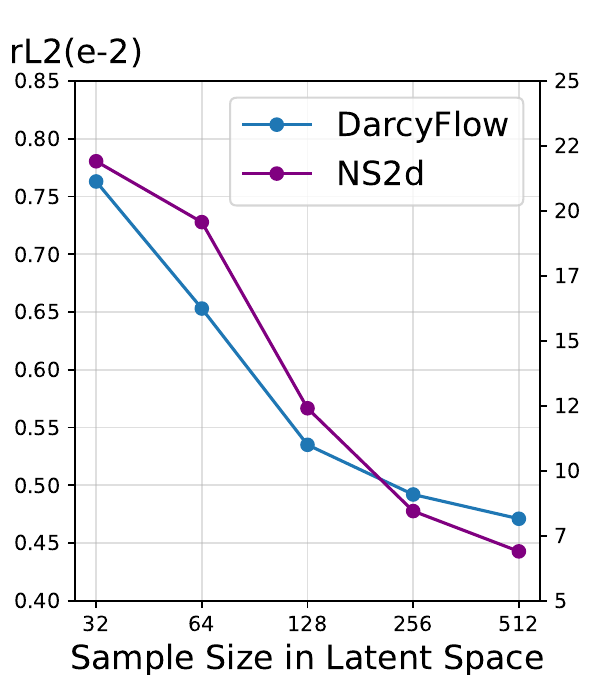}	
}
\noindent
\subfigure[]{
\label{fig-4c}
\includegraphics[width=0.31\linewidth]{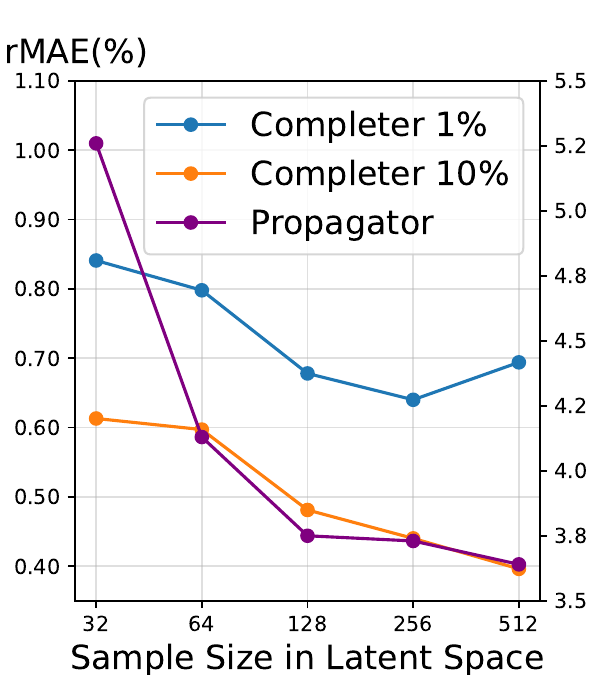}
}
\caption{Sampling numbers in latent space v.s. Accuracy. (a) Forward problems on Airfoil, Elasticity, Plasticity and Pipe. (b) Forward problems on Darcy and NS2d. (c) Inverse Problem.}
\label{fig-4}
\vspace{-10pt}
\end{figure}

\paragraph{Sampling numbers in latent space } We investigate the impact of hypothetical sampling position number in the latent space on LNO for solving both the forward and inverse problems. Results are shown in Figure~\ref{fig-4}. In the forward problems, as the sampling number $M$ in the latent space increases, the performance of LNO gradually improves to a saturation level, followed by a gentle decline on the benchmarks of Airfoil, Elasticity, Plasticity, and Pipe. However, on benchmarks of Darcy and NS2d, the performance of LNO continues to improve till $M=512$. This may be attributed to that the former set of benchmarks containing only location information, while the latter set involves both location and physics quantity information, which requires more sampling positions to represent. In the inverse problem, the performance of the completer and propagator are similar to the trends in the forward problems.

\begin{table}[htbp]
\centering
\caption{The ablation results using different attention implementations. P.A. stands for Physics-Attention. E.A. stands for Efficient Attention, N.A. stands for Nystrom Attention and G.A. stands for Galerkin-type Attention.}
\label{tab-4}
\begin{tabular}{lcccccc}
\toprule
\multirow{2}*{Model} & \multicolumn{6}{c}{Relative L2($\times 10^{-2}$)} \\
\cmidrule{2-7}
~ & Darcy & NS2d & Airfoil & Elasticity & Plasticity & Pipe\\
\midrule
LNO with P.A.\cite{Transolver}&0.53&22.65&0.74&0.88&0.49&0.40\\
LNO with E.A.\cite{EA,GNOT}&0.63&20.18&0.59&0.54&0.44&0.36\\
LNO with N.A.\cite{NA,ONO}&0.67&23.03&0.63&0.79&0.30&0.48\\
LNO with G.A.\cite{GalerkinTransformer}&0.71&22.44&0.89&0.93&0.43&0.39\\
LNO&\textbf{0.49}&\textbf{8.45}&\textbf{0.51}&\textbf{0.52}&\textbf{0.29}&\textbf{0.26}\\
\bottomrule
\end{tabular}
\end{table}

\begin{table}[htbp]
\centering
\caption{The ablation results of strategy of sharing weights on six forward problem benchmarks. Relative L2 is recorded. Smaller values represent better performance. The better result is in bold.}
\label{tab-3e}
\begin{tabular}{lcccccc}
\toprule
\multirow{2}*{Model} & \multicolumn{6}{c}{Relative L2($\times 10^{-2}$)} \\
\cmidrule{2-7}
~ & Darcy & NS2d & Airfoil & Elasticity & Plasticity & Pipe\\
\midrule
LNO(non-shared)&0.54&10.67&0.53&1.09&\textbf{0.26}&0.36\\
LNO&\textbf{0.49}&\textbf{8.45}&\textbf{0.51}&\textbf{0.52}&0.29&\textbf{0.26}\\
\bottomrule
\end{tabular}
\vspace{-10pt}
\end{table}

\section{Conclusions}
\label{conclusion}
The paper explores the feasibility of characterizing PDEs in a learnable latent space. We present \textbf{Ph}ysics-\textbf{C}ross-\textbf{A}ttension (PhCA) for encoding inputs into the latent space and for decoding outputs. We build the Latent Neural Operator (LNO) based on PhCA and validate its flexibility, efficiency and accuracy for solving both forward and inverse problems. We leave the generalization ability and re-usability of PhCA across multiple types of PDEs as another open research topic. 

Our work has also some limitations. For time-dependent equations, we decode the states at intermediate time steps into geometric space to compute the loss. Implementing additional training strategies may allow us to conduct all operations of the dynamic evolution process directly in the latent space, which potentially enhances efficiency. Also, in our work, no prior is imposed on the latent space. However, in the future, it is worth exploring whether physical priors, {\em e.g.,} symmetry, can help gain further precision improvement. 

\newpage
{\small
\bibliographystyle{unsrt}
\bibliography{ref1}
}

\newpage
\appendix

\section{Scaling}
\label{appendix-A}
We conduct scaling experiments to investigate the impact of model depth (number of stacked Transformer blocks) and model width (dimension of tokens in Transformer blocks) on the performance of LNO. 

\begin{figure}[htbp]
\centering
\scalebox{0.8}{
\includegraphics[width=1.0\textwidth]{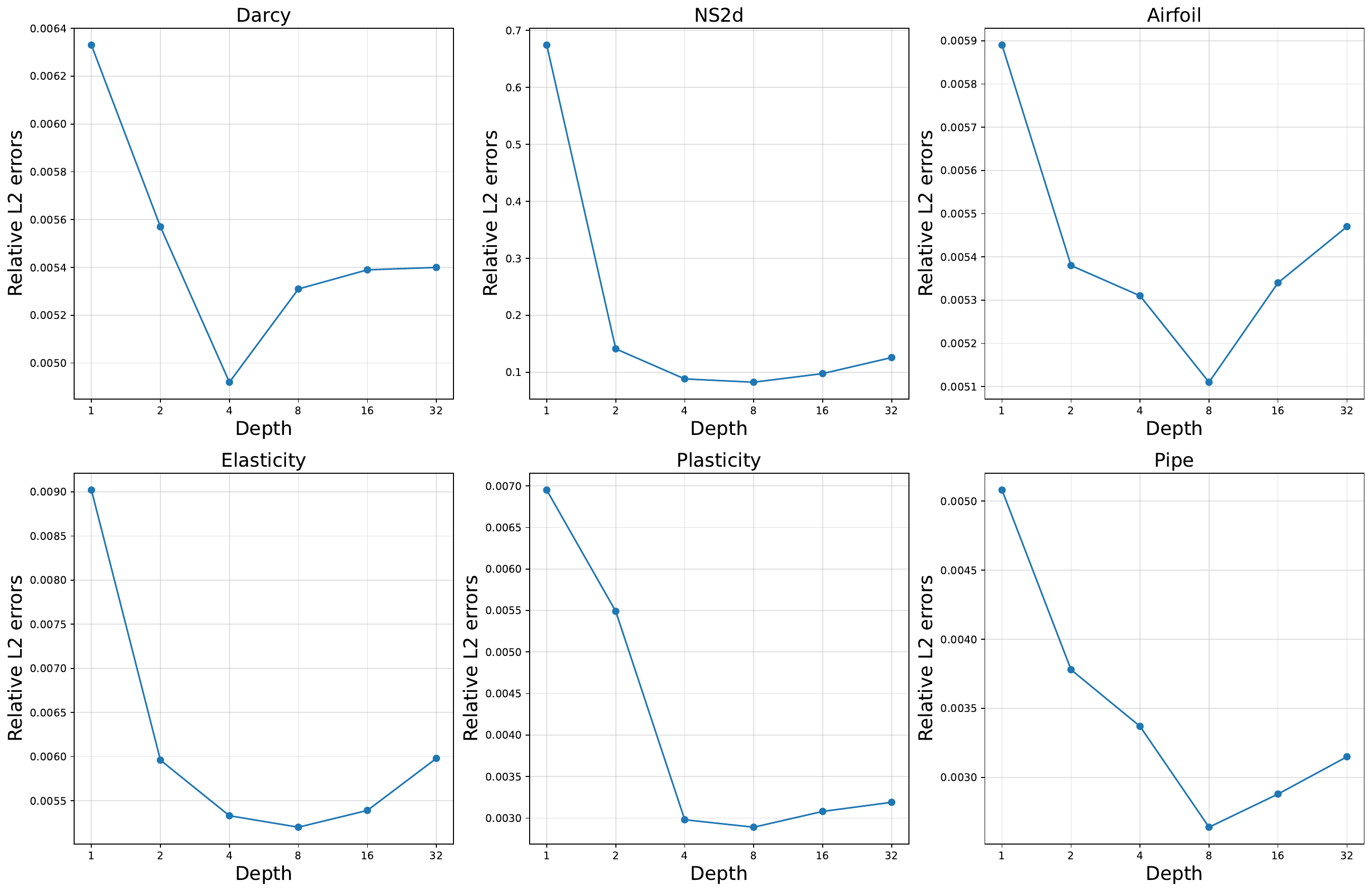}
}
\caption{The influence of model depth on the performance of LNO across six forward problem benchmarks.} 
\label{fig-a1}
\end{figure}

As shown in Figure~\ref{fig-a1}, with model depth increasing, the performance of LNO initially improves and then declines, reaching its optimal performance at 4 or 8. This suggests that deeper LNO encounters optimization challenges in solving PDEs in the latent space. Simply stacking more Transformer blocks does not necessarily lead to better accuracy. 

\begin{figure}[htbp]
\centering
\scalebox{0.8}{
\includegraphics[width=1.0\textwidth]{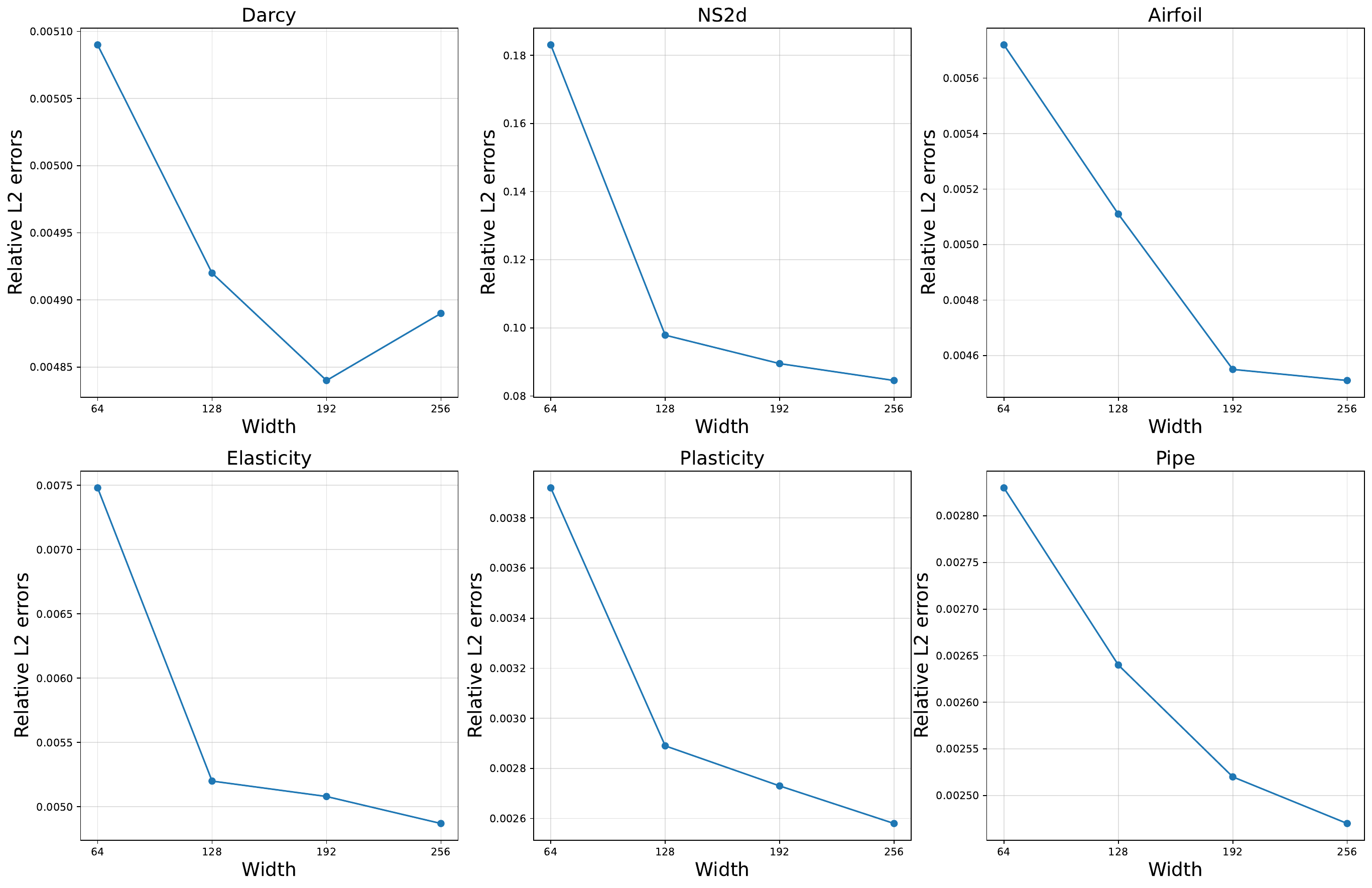}
}
\caption{The influence of model width on the performance of LNO across six forward problem benchmarks.} 
\label{fig-a2}
\end{figure}

As shown in Figure~\ref{fig-a2}, although the performance of LNO continues benefiting from increasing model width, it gradually exhibits a saturation trend when the dimension of tokens reaches as large as 256.

\section{Observation Ratio and Locations for the Inverse PDE Task}
\label{appendix-B}
\begin{figure}[htbp]
\centering
\scalebox{0.8}{
\subfigure[]{
\label{fig-b1}
\includegraphics[width=0.48\linewidth]{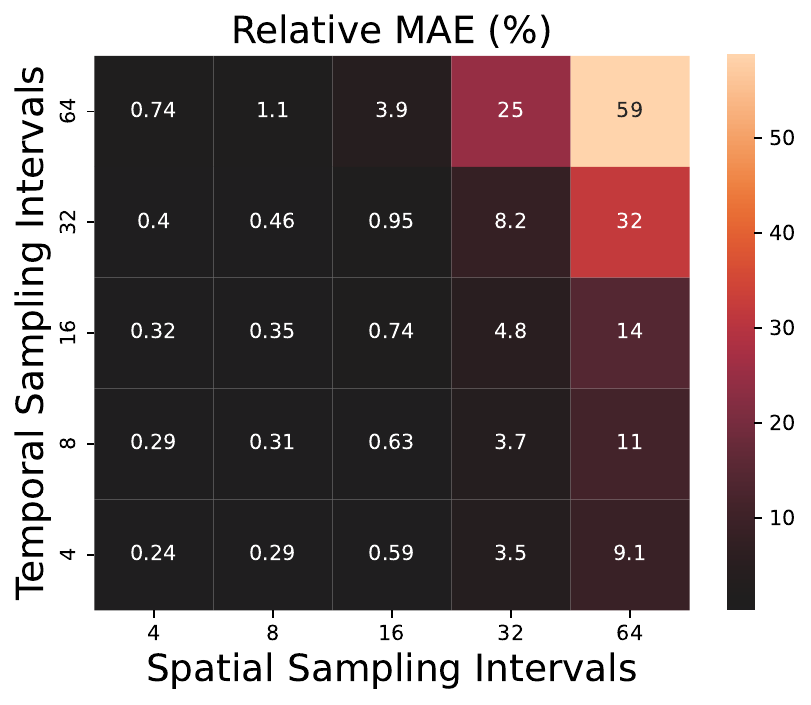}
}
\noindent
\subfigure[]{
\label{fig-b2}
\includegraphics[width=0.48\linewidth]{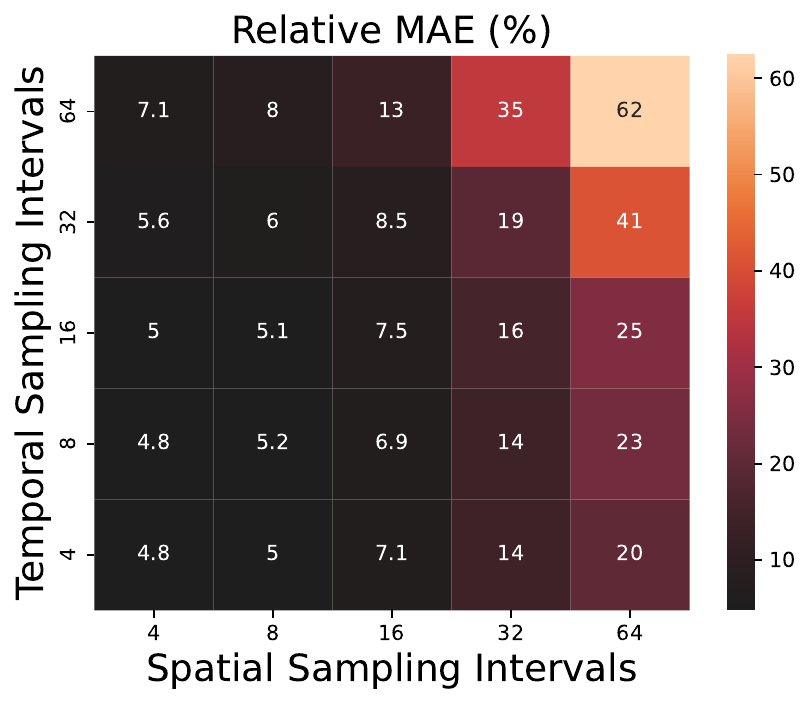}	
}}
\caption{Accuracy of solving the inverse problem at different temporal and spatial sampling intervals. (a) Accuracy of LNO as completer. (b) Accuracy of LNO as propagator based on the results of the corresponding completer.}
\label{fig-b}
\end{figure}

We investigate the impact of different temporal and spatial sampling intervals on the accuracy of LNO when solving inverse problem in the fixed observation situation. As shown by Figure~\ref{fig-b}, it can be found that i) the inverse problem can still be solved accurately even when the temporal and spatial sampling intervals are both as large as 16 (approximately only 0.4\% observation ratio); ii) the increase of the spatial sampling interval significantly compromises the solution accuracy compared to that of the temporal sampling interval; iii) spatial undersampling can hardly be compensated for through temporal oversampling.

\begin{figure}[htbp]
\centering
\includegraphics[width=0.8\textwidth]{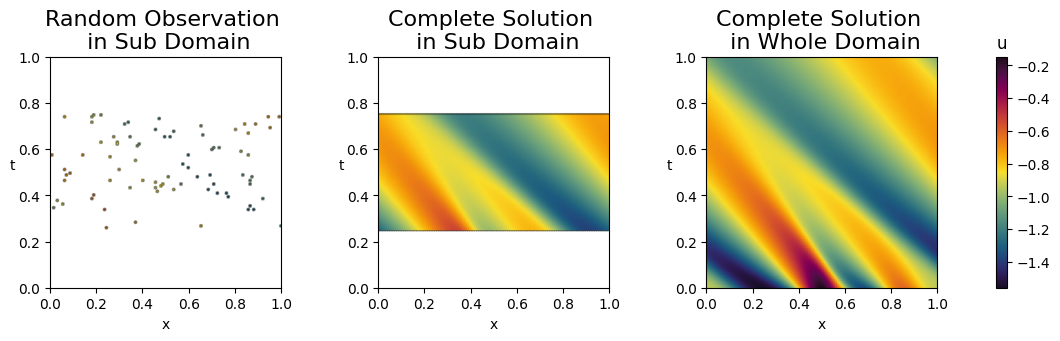}
\caption{Visualization of Burgers' equation with different observation situations in different regions. We propose a two-stage strategy. First, we interpolate the solution in the subdomain. Then, we extrapolate from the subdomain to the whole domain.} 
\label{fig-3}
\end{figure}  

\begin{figure}[htbp]
\centering
\includegraphics[width=0.8\textwidth]{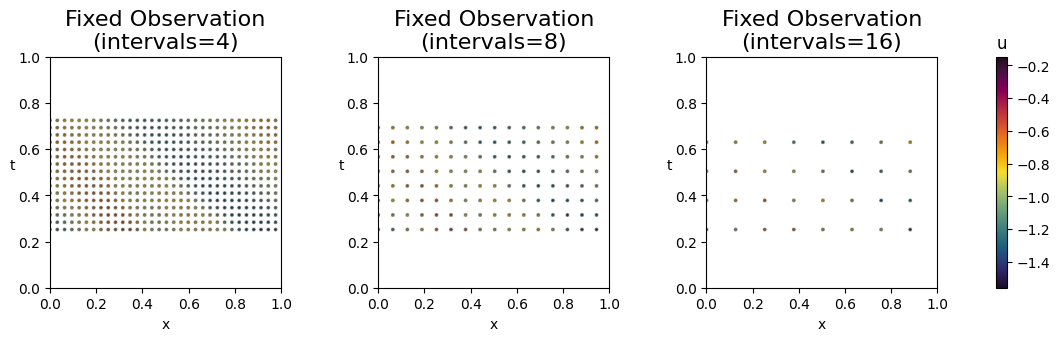}
\caption{Visualization of Burgers' equation in the fixed observation situation with different temporal and spatial sampling intervals.} 
\label{fig-bb}
\end{figure}

\section{Significance}
\label{appendix-C}
We conduct repeated experiments on both the forward and inverse problems, initializing the weights of our model and the reproduced models with different random seeds each time. We report the mean and standard deviation of these experiments to demonstrate the significance of our experimental results, as shown in Figure \ref{tab-c1} and Figure \ref{tab-c2}.

\begin{table}[htbp]
\centering
\caption{The mean and standard deviations (computed based on 5 independent trials) of different models in solving the forward problem.}
\label{tab-c1}
\scalebox{0.9}{
\begin{tabular}{lcccccc}
\toprule
\multirow{2}*{Model} & \multicolumn{6}{c}{Relative L2($\times 10^{-2}$)} \\
\cmidrule{2-7}
~&Darcy&NS2d&Airfoil&Elasticity&Plasticity&Pipe\\
\midrule
F-FNO\cite{FFNO}&0.75($\pm$0.01)&11.51($\pm$0.25)&0.60($\pm$0.03)&1.85($\pm$0.03)&0.27($\pm$0.02)&0.68($\pm$0.02)\\
U-FNO\cite{UFNO}&1.28($\pm$0.02)&17.15($\pm$0.29)&1.19($\pm$0.05)&2.13($\pm$0.03)&0.41($\pm$0.03)&0.62($\pm$0.02)\\
GNOT\cite{GNOT}&1.04($\pm$0.02)&13.40($\pm$0.28)&0.75($\pm$0.02)&0.88($\pm$0.03)&3.19($\pm$0.07)&0.45($\pm$0.02)\\
Transolver\cite{Transolver}&0.58($\pm$0.01)&8.79($\pm$0.17)&0.47($\pm$0.02)&0.62($\pm$0.02)&0.12($\pm$0.02)&0.31($\pm$0.01)\\
LNO(Ours)&0.49($\pm$0.01)&8.45($\pm$0.22)&0.51($\pm$0.05)&0.52($\pm$0.03)&0.29($\pm$0.03)&0.26($\pm$0.03)\\
\bottomrule
\end{tabular}
}
\end{table}

\begin{table}[htbp]
\centering
\caption{The mean and standard deviations (computed based on 3 independent trials) of different models in solving the inverse problem. Relative MAEs are expressed as percentages. The second to the second-to-last columns correspond to the situation where models act as completers under different observation rates, while the last column corresponds to the situation where models act as propagators.}
\label{tab-c2}
\scalebox{0.9}{
\begin{tabular}{lcccccc}
\toprule
Model&20\%&10\%&5\%&1\%&0.5\%&propagator\\
\midrule
DeepONet\cite{DeepONet}&2.37($\pm$0.18)&2.52($\pm$0.04)&2.77($\pm$0.10)&3.18($\pm$0.03)&4.70($\pm$0.31)&7.42($\pm$0.37)\\
GNOT\cite{GNOT}&1.23($\pm$0.07)&1.33($\pm$0.06)&1.63($\pm$0.04)&1.75($\pm$0.02)&2.53($\pm$0.08)&5.43($\pm$0.10)\\
LNO(Ours)&0.60($\pm$0.03)&0.73($\pm$0.11)&0.78($\pm$0.04)&1.26($\pm$0.09)&1.97($\pm$0.04)&3.46($\pm$0.21)\\
\bottomrule
\end{tabular}
}
\end{table}

\section{Full Implementation Details}
\label{appendix-D}
The full implementation details of LNO in both forward and inverse problems are shown in Table~\ref{tab-d1}.  All our experiments are conducted on single or two NVIDIA RTX 3090 GPUs. 

For the forward problems, we train the models using relative L2 error for 500 epochs. AdamW optimizer\cite{AdamW} and OneCycleLr scheduler\cite{OneCycleLR} are applied with initial learning rate at $10^{-3}$. We reproduce the results of Transolver\cite{Transolver} following implementations in appendix B.3 in \cite{Transolver} and also excerpt the results of other methods from Table 2 in \cite{Transolver}. The data splits are the same as in Appendix B.1 in \cite{Transolver}.

For the inverse problem, we train the models using mean squared error and StepLr scheduler. For DeepONet\cite{DeepONet}, we use 4 Transformer blocks with 128-dimensional tokens as the branch net and 3 fully-connected layers as the trunk net. For GNOT\cite{GNOT}, we set the number of Transformer blocks as 4, the dimension of tokens as 96, and the number of experts as 1. We use 4096 samples for training, 128 samples for validating and 128 samples for testing.

\begin{table}[htbp]
\centering
\caption{The hyperparameters and training configuration of LNO in solving the forward and inverse problems in Section \ref{forward-exp} and \ref{inverse-exp}}
\label{tab-d1}
\scalebox{0.85}{
\begin{tabular}{lccccccccc}
\toprule
Benchmark&Layers&Dim&Size&Heads&Batch&Epoch&Loss&Optimizer&Scheduler  \\
\midrule
Darcy&4&128&\multirow{6}*{256}&\multirow{6}*{8}&\multirow{6}*{4}&\multirow{6}*{500}&\multirow{6}*{rL2}&\multirow{6}*{AdamW\cite{AdamW}}&\multirow{6}*{OneCycleLR\cite{OneCycleLR}}\\
NS2d&8&256&~&~&~&~\\
Airfoil&8&128&~&~&~&~\\
Elasticity&8&128&~&~&~&~\\
Plasticity&8&128&~&~&~&~\\
Pipe&8&128&~&~&~&~\\
Completer&\multirow{2}*{4}&\multirow{2}*{96}&\multirow{2}*{256}&\multirow{2}*{8}&\multirow{2}*{4}&\multirow{2}*{500}&\multirow{2}*{MSE}&\multirow{2}*{AdamW}&\multirow{2}*{StepLR}\\
Propagator&~&~&~&~&~&~\\
\bottomrule
\end{tabular}
}
\end{table}

\end{document}